\documentclass{article}

    \usepackage[preprint, nonatbib]{neurips_data_2024}
\usepackage[utf8]{inputenc} % allow utf-8 input
\usepackage[T1]{fontenc}    % use 8-bit T1 fonts
\usepackage{hyperref}       % hyperlinks
\usepackage{url}            % simple URL typesetting
\usepackage{booktabs}       % professional-quality tables
\usepackage{amsfonts}       % blackboard math symbols
\usepackage{nicefrac}       % compact symbols for 1/2, etc.
\usepackage{microtype}      % microtypography
\usepackage{xcolor}         % colors
\usepackage{graphicx}
\usepackage{listings}
\usepackage{amsmath}
\usepackage[scaled]{beramono}
\usepackage{multirow}
\definecolor{eclipseStrings}{RGB}{42,0.0,255}
\definecolor{eclipseKeywords}{RGB}{127,0,85}
\definecolor{background}{HTML}{EEEEEE}
\colorlet{numb}{magenta!60!black}

\lstdefinelanguage{json}{
    basicstyle=\normalfont\ttfamily,
    commentstyle=\color{eclipseStrings}, % style of comment
    stringstyle=\color{eclipseKeywords}, % style of strings
    numbers=right,
    numberstyle=\scriptsize,
    stepnumber=1,
    numbersep=8pt,
    showstringspaces=false,
    breaklines=true,
    frame=lines,
    backgroundcolor=\color{background}, %only if you like
    string=[s]{"}{"},
    comment=[l]{:\ "},
    morecomment=[l]{:"},
    literate=
        *{0}{{{\color{numb}0}}}{1}
         {1}{{{\color{numb}1}}}{1}
         {2}{{{\color{numb}2}}}{1}
         {3}{{{\color{numb}3}}}{1}
         {4}{{{\color{numb}4}}}{1}
         {5}{{{\color{numb}5}}}{1}
         {6}{{{\color{numb}6}}}{1}
         {7}{{{\color{numb}7}}}{1}
         {8}{{{\color{numb}8}}}{1}
         {9}{{{\color{numb}9}}}{1}
}

\title{cPAPERS: A Dataset of Situated and Multimodal Interactive Conversations in Scientific Papers}

\author{%
  Anirudh Sundar$^{*}$ \quad   Jin Xu$^{*}$ \quad William Gay \quad Christopher Richardson \quad Larry Heck  \\
  AI Virtual Assistant Lab\\
  Georgia Institute of Technology\\
  $^{*}$ Equal Contribution \\
  \texttt{\{asundar34, jxu81, wgay7, crichardson8, larryheck\}@gatech.edu} \\
}

\begin{document}

\maketitle

\begin{abstract}
An emerging area of research in situated and multimodal interactive conversations (SIMMC) includes interactions in scientific papers. Since scientific papers are primarily composed of text, equations, figures, and tables, SIMMC methods must be developed specifically for each component to support the depth of inquiry and interactions required by research scientists.      This work introduces \textsc{Conversational Papers} (cPAPERS), a dataset of conversational question-answer pairs from reviews of academic papers grounded in these paper components and their associated references from scientific documents available on arXiv. We present a data collection strategy to collect these question-answer pairs from OpenReview and associate them with contextual information from \LaTeX source files. Additionally, we present a series of baseline approaches utilizing Large Language Models (LLMs) in both zero-shot and fine-tuned configurations to address the cPAPERS dataset. 
\end{abstract}

\section{Introduction}
Developing conversational assistants capable of situated and multimodal interactive conversations (SIMMC) over structured knowledge sources remains an open problem \cite{SIMMC:2020, sundar-heck-2022-multimodal,kottur-moon-2023-overview}. 

An emerging area of research within this domain is conversational interactions over scientific documents \cite{ramesh-etal-2023-comparative}. The number of scientific articles has grown dramatically over the past decade, making it difficult for scientists to find, read, understand, and connect advancements published by their fellow researchers. Scientists in the field of biomedicine, for example, publish over 1 million articles per year or a new article every 2 minutes on average \cite{landhuis2016scientific}. Developing methods to understand scientific documents and assist researchers is an important problem, especially for the Natural Language Processing (NLP) community. Scientific documents present an interesting challenge in SIMMC-based methods since the content is frequently multimodal \cite{blecher_nougat_2023}.

Besides textual paragraphs, researchers rely on various modalities to describe research methods. Figures convey information about concepts developed throughout the paper.  Generally, figures represent many different types of information including most commonly graphical and image information. For example, figures explain model architectures and pipelines, summarize experimental results in graphical plots, and represent complex information such as gradient learning surfaces and feature maps. In addition, figures can be used to show various stages of image processing in computer vision and image generation papers. 

Equations are crucial for grasping mathematical concepts in scientific texts but can be challenging to interpret. Equations frequently rely on notation introduced elsewhere in the document, often requiring readers to draw from the entire paper to understand the formulation. In addition, equations are typically related to and in many cases help the reader understand other paper components such as figures and tables (and vice-versa).  

Finally, tables typically have structures that convey semantic information. These include the text in the row and column headings, and text and often links in the table cells. These structures summarize multiple concepts (e.g., experimental results) in a human-readable form. Therefore, developing SIMMC methods for scientific papers must include a semantic understanding of these structures and table content.  

To advance the development of conversational assistants capable of SIMMC in scientific papers, this paper introduces Conversational Papers (\mbox{cPAPERS}\footnote{The dataset is publicly available at \url{https://huggingface.co/datasets/avalab/cPAPERS}. The collection process, baseline models, and code will be released with the camera-ready version under GNU Public License.}), a dataset of conversations in English situated in equations (\mbox{cPAPERS-EQNS}), figures (\mbox{cPAPERS-FIGS}), and tabular information (\mbox{cPAPERS-TBLS}) for scientific texts. Question-answer pairs are sourced from reviews and rebuttals from OpenReview~\footnote{\url{https://openreview.net/}}. Textual grounding to answer the questions and answers are sourced from the corresponding scientific articles hosted on arXiv~\footnote{\url{https://arxiv.org}}, an open-access repository of academic preprints. 

The contributions of this work include:

\begin{itemize}
    \item Introduction of the \textsc{Conversational Papers} (cPAPERS) dataset including three splits (EQNS, FIGS, TBLS)
    \item Development of a novel and scalable approach for collecting question-answer pairs from OpenReview and linking them with relevant contextual information from open access tex sources on arXiv
    \item Creation of baseline LLM approaches that  utilize weakly grounded multimodal context for dialogue responses
\end{itemize}

The rest of the paper is organized as follows: Section \ref{sec:related_work} discusses related work in language modeling and question-answering for equations, tables, and figures. Section \ref{sec:dataset} provides a description of \mbox{cPAPERS} and details the dataset collection process. Next Sections \ref{sec:approach} describe baseline approaches to address this dataset  
with a series of experiments utilizing zero-shot prompting and parameter-efficient fine-tuning. Section \ref{sec:Results} discusses the results, while Sections \ref{sec:conclusions} and \ref{sec:limitations} address the conclusion and limitations, respectively.

\begin{figure*}[htbp]
  \centering
  \includegraphics[width=1\textwidth]{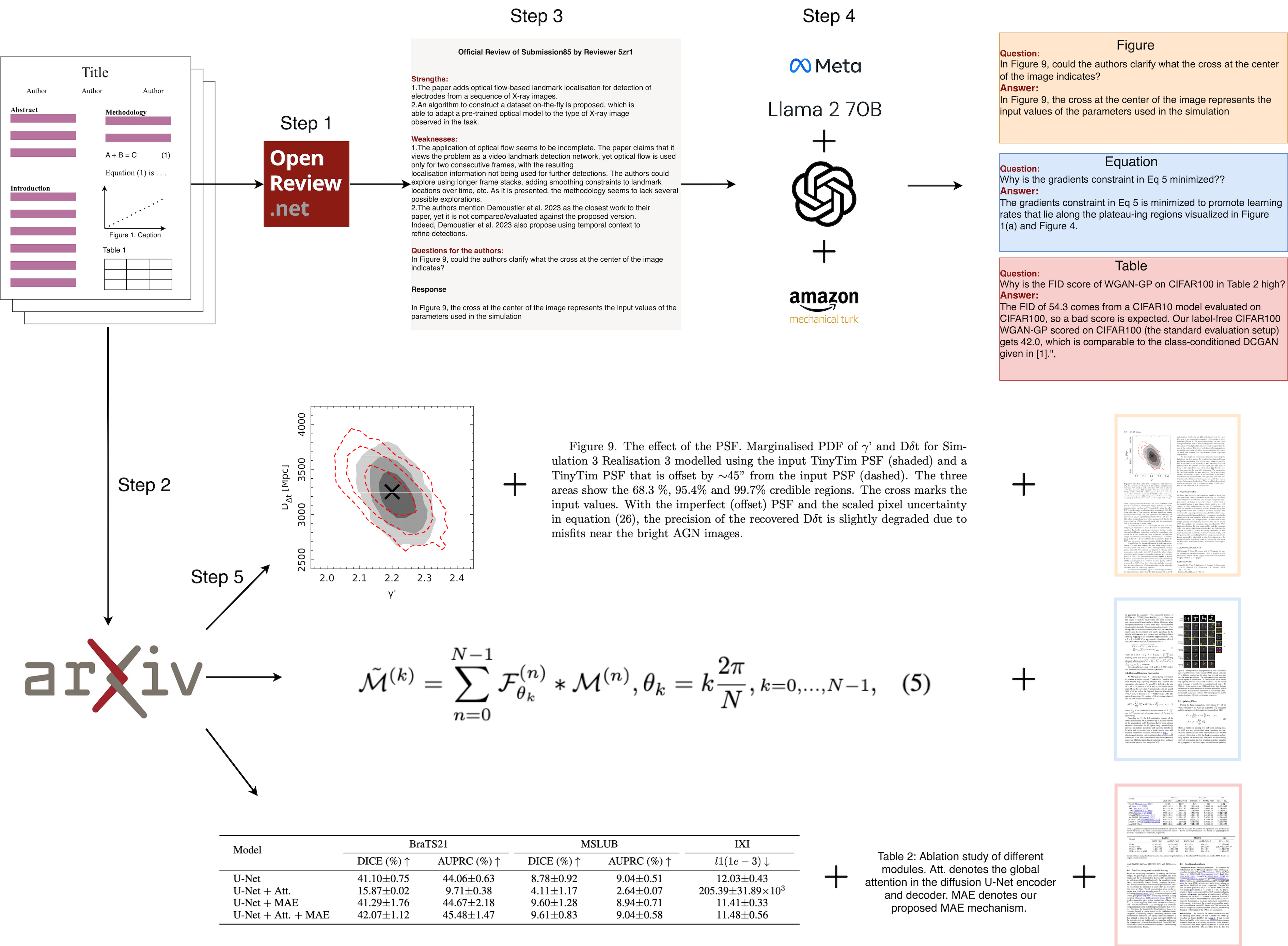}
  \caption{Overview of the data collection approach}
  \label{fig:overview}
\end{figure*}

\section{Related Work}
\label{sec:related_work}

\subsection{Language Modeling for Equations}

The challenge of modeling mathematical equations has become an active research area in Natural Language Processing. \cite{chiang-chen-2019-semantically} proposes an encoder-decoder approach to generate equations from math word problems. \cite{wang-etal-2021-math} presents an approach for the dual problem of generating math word problems consistent with equations. Prior work has also proposed learning representations from equations using the Transformer encoder %MathBERT
\cite{peng2021mathbert}, Tree-based encoders \cite{wang2021scientific}, and RNNs \cite{yasunaga2019topiceq}. More recent work utilizing generative architectures includes MathGPT \cite{scarlatos-lan-2023-tree}, an auto-regressive model based on GPT-2 \cite{radford2019language} for various language+equation tasks, and Nougat \cite{blecher_nougat_2023}, a Visual Transformer to parse academic documents to markup language. In contrast, \mbox{cPAPERS} is a dataset that addresses both grounded conversational question-answering and the modeling of mathematical equations. Answering the questions accurately requires understanding the expressions in the equations while also formulating an accurate response that addresses the task succinctly. 

\subsection{Question-Answering over Figures}
Early datasets on question-answering over figures include VQA \cite{agrawal_vqa_2016}, Visual7W \cite{zhu_visual7w_2016}, VisDial \cite{das_visual_2017}, \textsc{ManyModalQA} \cite{hannan_manymodalqa_2020}, and OK-VQA \cite{marino_ok-vqa_2019, reichman_outside_2023}. More recent datasets addressing the problem of open-domain conversations include IGC \cite{mostafazadeh_image-grounded_2017}, MOD \cite{fei_towards_2021}, and Image-Chat \cite{shuster_image_2020}. However, the images in these datasets are collected from MS-COCO \cite{lin_microsoft_2014} or YFCC100M \cite{thomee_yfcc100m_2016}, where visual content targets commonly seen everyday objects as opposed to scientific documents where images target specific information relevant to the explanation of a concept.

Recent work has addressed some of the challenges associated with multimodal image+text tasks situated in scientific documents. \cite{tan_scientific_2022} presents a dataset of charts from scientific papers and associated natural language captions summarizing the information present in the chart.  \cite{gong_recognizing_2021} develops a method to link labels with relevant images in patents. \cite{clark_extracting_nodate} introduces a dataset of 150 computer science papers, ground truth labels for the locations of figures, tables, and captions, and an approach to automatically extract this information from PDFs. 
\textsc{SciCap} \cite{hsu-etal-2021-scicap-generating} is a much larger dataset of 400,000 figures, their captions, and associated textual references from various scientific papers. The dataset is collected by scraping scientific preprints from arXiv. For each figure, they provide the associated caption and all paragraphs that mention the figure. 

While these datasets address tasks related to multimodal content in scientific documents, they are not conversational in nature. 
In contrast, \mbox{cPAPERS} addresses the shortcomings of prior datasets as a multimodal, conversational dataset grounded in scientific documents. The questions and answers address specific contextual visual information grounded in scientific documents. 

\subsection{Tabular Question-Answering}

Tabular question-answering addresses the problem of extractive QA grounded in the information contained in specific cells of a table. 
Prior tabular QA datasets include those collected from Wikipedia, such as \textsc{\mbox{WikiTableQuestions}} \cite{pasupat-liang-2015-compositional}, ManyModalQA \cite{hannan2020manymodalqa}, \textsc{TaBERT} \cite{yin-etal-2020-tabert},  NQ-Tables \cite{herzig-etal-2021-open}, FEVEROUS \cite{aly-etal-2021-fact}, FeTaQA \cite{nan-etal-2022-fetaqa}, \textsc{HybriDialogue} \cite{nakamura-etal-2022-hybridialogue}, and HiTab \cite{cheng-etal-2022-hitab}. Other tabular datasets are constructed from financial reports including TAT-QA \cite{zhu-etal-2021-tat}, \textsc{FinQA} \cite{chen-etal-2021-finqa}, \textsc{MultiHiertt} \cite{zhao-etal-2022-multihiertt}, or arXiv \mbox{iTBLS} \cite{sundar2024itbls}. 

Proposed approaches to address the tabular QA task include architectures based off of the Transformer encoder \cite{yin-etal-2020-tabert,herzig-etal-2020-tapas, Chen2020TabFact:, eisenschlos-etal-2020-understanding, liu2021tapex, gu-etal-2022-pasta, yang-etal-2022-tableformer}, decoder \cite{gong-etal-2020-tablegpt, akhtar-etal-2023-exploring,zha2023tablegpt,jiang-etal-2023-structgpt,zhang2023tablellama, 10.1145/3616855.3635752}, or both (encoder-decoder) \cite{nakamura-etal-2022-hybridialogue,deng-etal-2022-pacific, sundar-heck-2023-ctbls, sarkar2023testing, sundar2024itbls}. 

In contrast to prior tabular datasets, \mbox{cPAPERS-TBLS} presents a new source of grounded questions and answers and to the best of our knowledge, is the first to be situated in tables from OpenReview. Additionally, the question-answer pairs are not factoids. Rather, they are conversational in nature.

\begin{table*}[t]
\centering
\begin{tabular}{lccc | ccc | ccc}
\toprule
 & \multicolumn{3}{c}{Equation} & \multicolumn{3}{c}{Table} & \multicolumn{3}{c}{Figure} \\
 & train & dev & test & train & dev & test & train & dev & test \\
 \midrule 
\# Unique Papers & 672 & 286 & 335 & 715 & 285 & 302 & 761 & 275 & 313 \\
\# QA Pairs & 993 & 336 & 394 & 932 & 327 & 342 & 1052 & 297 & 357 \\
% \# Unique Papers & \multicolumn{3}{c}{1095} & \multicolumn{3}{c}{1471} & \multicolumn{3}{c}{714} \\
% \# QA Pairs & \multicolumn{3}{c}{1202} & \multicolumn{3}{c}{1722} & \multicolumn{3}{c}{1377} \\
\midrule 
\# Tokens (average) &  \multicolumn{3}{c}{ } &  \multicolumn{3}{c}{ } & \multicolumn{3}{c}{ }\\
            % \midrule 
            & train & dev & test & train & dev & test & train & dev & test \\
            \midrule 
Question & 25 & 25 & 25 & 24 & 22 & 26 & 23 & 23 & 24  \\
Answer & 92 & 102 & 90 & 86 & 79 & 81 & 83 & 88 & 81 \\
Contexts & 10,232 & 11,851 & 12,288 & 2,981 & 2,746 & 2,610 & 433 & 400 & 431 \\
References & 7,323 & 8,413 & 9,517 & 1,757 & 1,645 & 1,427 & 366 & 375 & 323 \\
Neighboring Contexts & 1,144 & 1,153 & 1,084 & 994 & 1,043 & 925 & - & - & - \\
Neighboring References & 1,000 & 947 & 1,152 & 736 & 657 & 588 & - & - & - \\
\bottomrule
\end{tabular}
\caption{Dataset Statistics}
\label{tab:dataset_summary}
\end{table*}

\section{The cPAPERS Dataset}
\label{sec:dataset}

\subsection{Dataset Description}
Conversational Papers (\mbox{cPAPERS}) is a dataset of conversations in English situated in scientific texts. \mbox{cPAPERS} consists of question-answer pairs pertaining to figures (\mbox{cPAPERS-FIGS}), equations (\mbox{cPAPERS-EQNS}), or tabular information (\mbox{cPAPERS-TBLS}) from scientific papers. These pairs are sourced from the official reviews and rebuttals of The Conference and Workshop on Neural Information Processing Systems (NeurIPS) and The International Conference on Learning Representations (ICLR) between 2020 and 2023, as available on OpenReview. \mbox{cPAPERS} comprises of 1723 question-answer pairs over equations, 1601 pairs over tables, and 1706 pairs over figures, totaling 5030 question-answer pairs spanning 2350 unique scientific papers. A detailed breakdown of the dataset is provided in Table \ref{tab:dataset_summary}.

Section \ref{subsec:dataset_collection} details the steps of collecting \mbox{cPAPERS} from official reviews on OpenReview, extracting question-answer pairs, and associating them with \LaTeX source files from arXiv.

\subsection{Dataset Collection}
\label{subsec:dataset_collection}
\subsubsection{Step 1 - Extract Official Review and Comments}

We leverage the OpenReview API to access official reviews and comments for each paper submission. These reviews, authored by conference reviewers, typically include a paper summary, strengths and weaknesses, specific questions, and limitations. Authors respond to these critiques in the comments, providing answers that address the reviewers' questions. This step involves downloading the official reviews and comments for all submissions.

\subsubsection{Step 2 - Download \LaTeX files}

For each paper recorded in Step 1, we initiate a call to the arXiv API to verify the existence of the paper. If the paper is found, we proceed to download all associated \LaTeX source files. If a paper cannot be located using the arXiv API, we exclude this submission from further processing.

\subsubsection{Step 3 - Regex Filtering}
As mentioned in Step 1, review-rebuttal pairs contain a summary of the paper and additional information that may not be related to equations, tables, or figures. Therefore, we use regular expressions to identify mentions of equations, tables, and figures within each review and comment pair. The following regular expressions are employed for this purpose:

\begin{verbatim}
Equation: re.compile(r'\b(?:Equation|Eq\.?)\s*\(?\s*\d+\)?\b', re.IGNORECASE)
\end{verbatim}

\begin{verbatim}
Table: re.compile(r'\bTable\s*\d+|\btables\b',re.IGNORECASE)
\end{verbatim}

\begin{verbatim}
Figure: re.compile(r'\b(?:Figure|fig)\.?\s*\d+\b', re.IGNORECASE)
\end{verbatim}

These expressions efficiently determine whether equations, tables, and figures are mentioned in either the review or the rebuttal, resulting in a separate list of review-rebuttal pairs each containing explicit references to an equation, table, or figure in the paper. It is possible for a review to contain references to multiple modalities.   

\subsubsection{Step 4 - Question-Answer Extraction}

To extract question-answer pairs from the regex-filtered reviews and rebuttals, an LLM that is instruction fine-tuned to align with human preferences is prompted \textit{in-context} with a single example. For this process, we use \textsc{Llama-2-70b-chat-hf} \cite{touvron2023llama}.

The prompt is as follows:

% \textbf{Prompt for Extracting QA Pairs}
\textit{`Given this example Content, Question, and Answer, look through the Summary and Comments that are dedicated to the paper and extract Question and Answer pairs that specifically target a particular Figure/Equation/Table in the content. Questions may be under sections labeled similarly to ``Weaknesses" or ``Questions” or ``Review"  or ``Clarity, Quality, Novelty And Reproducibility" in the comments, and Answers may be under sections labeled ``Response to Reviewer" or ``Rebuttal" or ``Comments".'}

System Prompt: \textit{``You are a helpful question answer extracting assistant. Our goal is to extract Question and Answer pairs pertaining to a specific a Figure/Equation/Table in the content. Your response should be in the format of [Question] <question> [Answer] <answer> Do not add any other unnecessary content in your response."}

\textbf{Post-processing + JSON reformatting.}
GPT 3.5 \cite{brown2020language} is used to reformat the question-answer pairs as JSON. The following system prompt is used.

Prompt: \textit{You are a helpful assistant. Please find the Question Answer pairs, and format it as a json (\{``question\_answers": [\{ ``question": "", ``answer": "" \},...]\})}

For each question-answer pair, an additional regular expression is employed to extract the respective figure, equation, or table number following the keywords `Figure', `Fig', `Equation', `Eq', or `Table'.

% \textcolor{red}{@Jin mention the prompt here too}
\begin{verbatim}
pattern = r'\d+'        
numbers = re.findall(pattern, input_string)
\end{verbatim}

% We observed that \textsc{Llama-2-70B} often repeats the example provided in the prompt when it fails to extract a question-answer pair. To address this issue, we implement an additional step to clean the resulting data by removing question-answer pairs that repeat or closely resemble the examples provided in the prompts. 

\subsubsection{Step 5 - Crowdworker Cleanup}
A significant portion of academic reviews and rebuttals on OpenReview pertain to clarification questions and fixing typos. While the LLM processing removes most of these spurious questions and answers, to further ensure the quality of the dataset we employ crowd workers from Amazon Mechanical Turk to ascertain whether a question-answer pair about an equation, table, or figure is technical in nature or asks to fix a typo. For each question-answer pair, two crowdworkers provide feedback on whether the question is technical in nature or not and only those question-answer pairs with a consensus are retained. 

\subsubsection{Step 6 - Contextualizing QA Pairs}
\label{sec:step_6}
In addition to extracting question-answer pairs from OpenReview, \mbox{cPAPERS} associates these question-answer pairs with referring text from the \LaTeX source on arXiv. 
Often, the text surrounding figures, equations, and tables in scientific documents provides additional clarifying information. Regular expressions are employed to locate all instances of figures, equation, and tables in the \LaTeX source using the \verb|\begin{figure}|, \verb|\begin{equation}|, and \verb|\begin{table}| environments.

\textbf{Equation:} The equation is obtained by extracting content enclosed between \verb|\begin{equation}| and \verb|\end{equation}| tags. We provide all equations from the \LaTeX source for a question-answer pair.

\textbf{Table:} The table is extracted from the content enclosed between \verb|\begin{tabular}| and \verb|\end{tabular}| tags. We provide all tables from the \LaTeX source for a question-answer pair.

\textbf{Figure:} First, the content between \verb|\begin{figure}| and \verb|\end{figure}| tags is extracted. From this content, we obtain the caption (using \verb|\caption|) and the figure path (using \verb|\includegraphics|). 

% \textbf{Equation/Table/Figure number:} We report the equation/table/figure number as mentioned in the review-rebuttal pair, determined by a regular expression. \textcolor{red}{@Jin mention the regex used to find equation/table/figure number}

Additionally, for each modality, we provide context and references:

\textbf{Context:} The context represents the paragraph of text preceding and the paragraph of text following the equation, table, or figure referred to in the question-answer pair. 

\textbf{References:} In addition, we report all referring text in a document associated with the equation, table, or figure in the question-answer pair. Within each environment, we search for a label (\verb|\label|) and if it exists, we locate all references by searching for text using the label in \LaTeX's \verb|\ref| command.

Reviews of academic pre-prints provide a source for high-quality document-grounded questions and answers and success on the task requires contextual and multimodal understanding. However, one of the main challenges with matching questions and answers from OpenReview with their corresponding multimodal context from arXiv is that while reviewer and author references to equations, tables, and figures are purely ordinal (e.g. Equation 3, Figure 4, Table 2), the \LaTeX source does not provide any reliable approach to associate this ordinal position with the multimodal context in the final compiled document, specifically for equations and tables. Additionally, arXiv contains multiple versions of papers with no guarantees on consistency with the submission to OpenReview. As a result, the \mbox{cPAPERS} dataset reports both the position of the equation, table, or figure referred to in the question, and all equations and tables from that specific paper. Section \ref{sec:approach} contains additional details on how this information is used in a baseline approach.

% \subsubsection{Step 6 - Contextualizing QA Pairs}

\subsubsection{Step 7 - Additional Post-processing of Figures}

Authors frequently utilize different graphical formats in their papers. To ensure consistency in our dataset, we convert all \verb|.pdf| and \verb|.eps| files to \verb|.png| format using ImageMagick \footnote{\url{https://imagemagick.org/}}.

\section{Baseline Approaches}
\label{sec:approach}

We experiment with zero-shot prompting and parameter-efficient fine-tuning for the baseline approaches. Zero-shot prompting involves querying pre-trained LLMs to answer the questions in the \mbox{cPAPERS} dataset without additional fine-tuning. In this approach, we use \textit{off-the-shelf} Transformer decoder LLMs already pre-trained using a causal language modeling objective followed by supervised fine-tuning and reinforcement learning with human feedback to follow instructions. The LLM is not fine-tuned on \mbox{cPAPERS} and we do not provide examples in-context. 

In addition, we experiment with parameter-efficient fine-tuning using QLoRA \cite{dettmers_qlora_2023} to better align the model with the style of responses in the \mbox{cPAPERS} dataset. An LLM is fine-tuned independently for each modality (equation, table, or figure). 

We serialize equations and tables using the corresponding \LaTeX representations from the \textsf{.tex} source on arXiv. % For figures, we utilize two settings - the first one uses LLMs on textual modalities (figure caption, surrounding context from \LaTeX, and referring text) while the second one uses a Large Multimodal Model pre-trained for vision+language tasks to use the actual image instead of the figure's caption. 
As previously indicated in Section \ref{sec:step_6}, pairing questions and answers from OpenReview with \LaTeX sources from arXiv presents a challenge with possible inconsistencies between the references in questions and answers to the corresponding sources in \LaTeX. Therefore, we conduct question-answering experiments under two settings - utilizing all multimodal content, or utilizing a smaller subset of neighboring weakly-grounded multimodal content. While using all multimodal content (equations or tables) ensures that the correct equation (or table) referred to by the question-answer pair is provided to the model, limitations on context length constrain the model from accurately using this information. Evidenced by prior work demonstrating the retrieval capabilities of generative LLMs when provided weakly-grounded context \cite{sundar-heck-2023-ctbls}, we reconcile this situation by providing the model with a subset of equations or tables from the entire paper. First, we assign each equation or table with an ordinal position based on  each instance of a \verb|\begin{equation}| or \verb|\begin{table}| in the \LaTeX source. Then, supposing the question-answer pair refers to equation$_i$, the LLM is provided a subset of equations: $\{\text{equation}_{i-1}, \text{equation}_{i}, \text{equation}_{i+1}\}$. We utilize one-sided grounding in the situation where grounding from either boundary does not exist. 

\section{Results}
\label{sec:Results}

The following sections contain results of the baseline approaches on the development and test splits. We report results from automatic evaluation - ROUGE \cite{lin_rouge_2004} \footnote{\url{https://huggingface.co/spaces/evaluate-metric/rouge}}, METEOR \cite{banarjee2005} \footnote{\url{https://huggingface.co/spaces/evaluate-metric/meteor}}, and BERTScore \footnote{\url{https://huggingface.co/spaces/evaluate-metric/bertscore}} \cite{zhang_bertscore_2020}. 

% \subsection{Zero-shot}

\textbf{Zero-shot.} First, we experiment with zero-shot language modeling with \textsc{Llama-2-70B} \cite{touvron2023llama} to answer questions on \mbox{cPAPERS-EQNS}, \mbox{cPAPERS-TBLS}, and \mbox{cPAPERS-FIGS}. The results from Tables \ref{tab:zs_eqn} and \ref{tab:zs_tab} indicate that utilizing the neighboring equations/tables provides a significant improvement over using all equations/tables to answer questions in the dataset, evidenced by up to 2x improvement in ROUGE, METEOR, and BERTScores. Additionally, the results from Table~\ref{tab:zs_eqn} indicate that utilizing the question alone without neighboring textual context (paragraphs of text surrounding the equation) results in best ROUGE and METEOR scores on \mbox{cPAPERS-EQNS}, whereas utilizing referring text yields the best scores on these metrics in \mbox{cPAPERS-TBLS} when compared to using the question alone to generate answers (Table \ref{tab:zs_tab}). 
%\begin{color}{red} CHECK THE NEXT SENTENCE FOR CORRECTNESS.  
We posit this is because  referring text in the document is predominantly used to summarize tabular results.

\begin{table*}[t]
\centering
\resizebox{\textwidth}{!}{\begin{tabular}{c|c|c|c|c|c|c}
\toprule 
\textbf{Modality} & \textbf{Setting} & \textbf{ROUGE-1} & \textbf{ROUGE-2} & \textbf{ROUGE-L} & \textbf{METEOR} & \textbf{BERTScore} \\ \midrule
Question (Q) & - & \textbf{0.194} & \textbf{0.065} & \textbf{0.144} & 0.240 & \textbf{0.825}\\ \midrule
\multirow{2}{*}{Q+Equation} & Neighboring & 0.190 & 0.063 & 0.139 & \textbf{0.245} & 0.821\\ 
& All & 0.170 & 0.056 & 0.123 & 0.218 & 0.740\\ \midrule
\multirow{2}{*}{Q+Context} & Neighboring & 0.186 & 0.063 & 0.137 & 0.237 & 0.809\\ 
& All & 0.079 & 0.027 & 0.058 & 0.102 & 0.345\\ \midrule
\multirow{2}{*}{Q+References} & Neighboring & 0.176 & 0.061 & 0.129 & 0.223 & 0.764\\ 
& All & 0.112 & 0.037 & 0.082 & 0.143 & 0.498\\ \bottomrule
\end{tabular}}
\caption{Zero-shot language modeling on \mbox{cPAPERS-EQNS} test set with \textsc{Llama-2-70B}}
\label{tab:zs_eqn}
\end{table*}

\begin{table*}[h]
\centering
\resizebox{\textwidth}{!}{\begin{tabular}{c|c|c|c|c|c|c}
\toprule 
\textbf{Modality} & \textbf{Setting} & \textbf{ROUGE-1} & \textbf{ROUGE-2} & \textbf{ROUGE-L} & \textbf{METEOR} & \textbf{BERTScore} \\ \midrule
Question (Q) & - & 0.192 & 0.058 & 0.136 & 0.232 & \textbf{0.832}\\ \midrule
\multirow{2}{*}{Q+Table} & Neighboring & 0.206 & 0.061 & \textbf{0.145} & 0.237 & 0.828\\ 
& All & 0.176 & 0.052 & 0.123 & 0.199 & 0.697\\ \midrule
\multirow{2}{*}{Q+Context} & Neighboring & 0.202 & 0.062 & 0.142 & 0.241 & 0.828\\ 
& All & 0.160 & 0.050 & 0.114 & 0.195 & 0.666\\ \midrule
\multirow{2}{*}{Q+References} & Neighboring & \textbf{0.207} & \textbf{0.064} & 0.144 & \textbf{0.243} & 0.829\\ 
& All & 0.186 & 0.057 & 0.130 & 0.223 & 0.758\\ \bottomrule
\end{tabular}}
\caption{Zero-shot language modeling on \mbox{cPAPERS-TBLS} test set with \textsc{Llama-2-70B}}
\label{tab:zs_tab}
\end{table*}

Table \ref{tab:zs_fig} contains results for zero-shot language modeling on \mbox{cPAPERS-FIGS}. In this situation, utilizing context performs the best across metrics.

\begin{table*}[h]
\centering
\begin{tabular}{c|c|c|c|c|c}
\toprule 
\textbf{Modality} & \textbf{ROUGE-1} & \textbf{ROUGE-2} & \textbf{ROUGE-L} & \textbf{METEOR} & \textbf{BERTScore} \\ \midrule 
 Question (Q)         & 0.185       & 0.065            & 0.137                & 0.238  & 0.833          \\ 
Q+Caption      & 0.200       & 0.074            & 0.149                & 0.248           & 0.837       \\ 
Q+Context      & \textbf{0.208}       & \textbf{0.076}            & \textbf{0.155}                & \textbf{0.254}           & \textbf{0.837}        \\ 
Q+References   & 0.205       & 0.075            & 0.154                & 0.251           & 0.837          \\ 
\bottomrule 
\end{tabular}
\caption{Zero-shot language modeling on \mbox{cPAPERS-FIGS} test set with \textsc{Llama-2-70B}}
\label{tab:zs_fig}
\end{table*}

% \subsection{Fine-tune}
\textbf{Fine-tuning}. We also report results by fine-tuning \textsc{Llama-2-7B} using QLoRA \cite{dettmers_qlora_2023}. Motivated by the zero-shot  xperiments, we only conduct fine-tuning using the \textit{neighboring} setting for \mbox{cPAPERS-EQNS} and \mbox{-TBLS}. 

Results of fine-tuning on \mbox{cPAPERS-EQNS} are detailed in Table \ref{tab:ft_eqn}. Utilizing the question with the equation outperforms the question alone or using the question along with referring text or the context on ROUGE-1 and ROUGE-L scores, METEOR, and BERTScore. However, the ANOVA test shows that the difference is not statistically significant.

\begin{table*}[t]
    \centering
    \begin{tabular}{*{5}{c|}c}
    \toprule 
    \textbf{Modality} & \textbf{ROUGE-1} & \textbf{ROUGE-2} & \textbf{ROUGE-L} & \textbf{METEOR} & \textbf{BERTScore} \\
    \midrule
    Question (Q) & 0.309 & \textbf{0.138} & 0.248 & 0.215 & 0.860 \\
    Q + Equation & \textbf{0.317} & 0.134 & \textbf{0.251} & \textbf{0.223} & \textbf{0.861} \\
    Q + Context & 0.297 & 0.126 & 0.235 & 0.221 & 0.817 \\
    Q + References & 0.283 & 0.122 & 0.224 & 0.217 & 0.777 \\
    \bottomrule
    \end{tabular}
    \caption{Results of fine-tuning \textsc{Llama-2-7B} on \mbox{cPAPERS-EQNS} test set}
    \label{tab:ft_eqn}
\end{table*}

In contrast, the results from finetuning \mbox{cPAPERS-TBLS} and \mbox{cPAPERS-FIGS} in Tables \ref{tab:ft_tab} and \ref{tab:ft_fig} indicate that using just the question results in the best performance across most metrics. We believe that this is a result of the nature of questions and answers in reviews where authors often utilize tables and figures to supplement reviewer questions. This is supported by Table \ref{tab:qa_comparison}. In the \mbox{cPAPERS-EQNS} split of the dataset, 55.9\% of questions refer to an equation while 77.6\% of answers refer to an equation. On \mbox{cPAPERS-TBLS} and \mbox{cPAPERS-FIGS}, this difference is exacerbated, with answers referring to a table or a figure more often and questions referring to them less often when compared to equations. 

\begin{table*}[h!]
    \centering
    \begin{tabular}{*{5}{c|}c}
    \toprule 
         \textbf{Modality} & \textbf{ROUGE-1} & \textbf{ROUGE-2} & \textbf{ROUGE-L} & \textbf{METEOR} & \textbf{BERTScore} \\
         \midrule 
         Question (Q) & \textbf{0.315} & \textbf{0.121} & \textbf{0.235} & 0.218 & \textbf{0.869} \\
         Q + Table & 0.293 & 0.107 & 0.218 & 0.212 & 0.820 \\
         Q + Context & 0.294 & 0.106 & 0.216 & \textbf{0.225} & 0.838 \\
         Q + References & 0.292 & 0.106 & 0.214 & 0.218 & 0.816 \\
         \bottomrule 
    \end{tabular}
    \caption{Results of fine-tuning \textsc{Llama-2-7B} on \mbox{cPAPERS-TBLS} test set}
    \label{tab:ft_tab}
\end{table*}

\begin{table*}[h!]
    \centering
    \begin{tabular}{*{5}{c|}c}
    \toprule 
         \textbf{Modality} & \textbf{ROUGE-1} & \textbf{ROUGE-2} & \textbf{ROUGE-L} & \textbf{METEOR} & \textbf{BERTScore} \\
         \midrule 
         Question (Q)     & \textbf{0.329} & \textbf{0.155} & \textbf{0.269} & \textbf{0.250} & 0.859 \\
         Q + Figure       & 0.322 & 0.147 & 0.260 & 0.246 & \textbf{0.868} \\
         Q + Context      & 0.321 & 0.140 & 0.256 & \textbf{0.250} & 0.858 \\
         Q + References   & 0.311 & 0.131 & 0.244 & 0.247 & 0.843 \\
         \bottomrule 
    \end{tabular}
    \caption{Results of fine-tuning \textsc{Llama-2-7B} on \mbox{cPAPERS-FIGS} test set}
    \label{tab:ft_fig}
\end{table*}

\begin{table}[h!]
    \centering
    \begin{tabular}{c|c|c}
    \toprule 
    \textbf{Split} & \textbf{Question (\%)} & \textbf{Answer (\%)} \\
    \midrule 
    \mbox{cPAPERS-EQNS} & 55.94 & 77.65 \\
    \mbox{cPAPERS-TBLS} & 46.47 & 79.51 \\
    \mbox{cPAPERS-FIGS} & 54.16 & 80.48 \\
    \bottomrule
    \end{tabular}
    \caption{Comparison between the percentage of questions that contain a reference to an Equation, Table, or Figure versus the percentage of answers }
    \label{tab:qa_comparison}
\end{table}

\section{Conclusion} 
\label{sec:conclusions}
This paper introduces the \textsc{cPAPERS} dataset, which addresses the shortcomings of prior datasets as a multimodal, conversational dataset grounded in scientific documents. Leveraging reviews of academic papers grounded in equations, figures, and tables, as well as their associated references from scientific documents available on arXiv, this dataset aims to further advance the development of conversational assistants capable of situated and multimodal interactive conversation within scientific papers.
We conduct a series of experiments with zero-shot prompting to benchmark state-of-the-art models performance and perform several parameter-efficient fine-tuning for the baseline approaches. The experimental results provide valuable insights and create opportunities for enhancing and innovating the development of future conversational AI assistant.

 % The fine-tuning experiments demonstrate improvements in model performance when utilizing additional contextual information along with the question. This opens up opportunit for for improvement and innovation in   Importantly, \textsc{cPAPERS} 

% \section{Limitations}
\section{Discussion}
\label{sec:limitations}
This paper introduces an innovative and scalable approach for collecting a large dataset of situated and multimodal interactive conversations related to scientific papers. This approach has the potential to facilitate the collection of extensive question-answer pairs across a wide range of scientific fields with ease. Collecting such a large and diverse dataset would benefit the research community by providing opportunities to develop more advanced AI-based research assistants for scientific research.

There may be concerns that creating an AI assistant to help automate the scientific discovery process could diminish the role of human scientists. While this may be a possible use of this dataset, our intention is to develop an AI research assistant that amplifies human scientists, empowering them to achieve greater scientific discoveries more efficiently.

\textbf{Limitations.} The key limitation of this dataset is the presence of mismatched figures, tables, or equations across different versions of the manuscripts. The \textit{.tex} files of the paper on arXiv often undergo multiple revisions, and  comments on Open Review are typically specific to a particular version. Authors frequently make changes in response to reviewer comments, which may involve additions, removals, or reordering of figures, equations, or tables. 
% \begin{color}{red} CAN WE QUANTIFY/ESTIMATE HOW BIG OF A PROBLEM THIS IS? \end{color} 
This presents a potential mismatch between the figures, equations, or tables and the question-and-answer pairs, thereby introducing additional challenges for language modeling.

\begin{ack}
This work was supported by NSF IIS-2112633 and by CoCoSys, one of seven centers in JUMP 2.0, a Semiconductor Research Corporation (SRC) program sponsored by DARPA.
\end{ack}

\bibliographystyle{plain}

\begin{thebibliography}{10}

\bibitem{agrawal_vqa_2016}
Aishwarya Agrawal, Jiasen Lu, Stanislaw Antol, Margaret Mitchell, C.~Lawrence Zitnick, Dhruv Batra, and Devi Parikh.
\newblock {VQA}: {Visual} {Question} {Answering}.
\newblock {\em arXiv:1505.00468 [cs]}, October 2016.
\newblock arXiv: 1505.00468.

\bibitem{akhtar-etal-2023-exploring}
Mubashara Akhtar, Abhilash Shankarampeta, Vivek Gupta, Arpit Patil, Oana Cocarascu, and Elena Simperl.
\newblock Exploring the numerical reasoning capabilities of language models: A comprehensive analysis on tabular data.
\newblock In Houda Bouamor, Juan Pino, and Kalika Bali, editors, {\em Findings of the Association for Computational Linguistics: EMNLP 2023}, pages 15391--15405, Singapore, December 2023. Association for Computational Linguistics.

\bibitem{aly-etal-2021-fact}
Rami Aly, Zhijiang Guo, Michael~Sejr Schlichtkrull, James Thorne, Andreas Vlachos, Christos Christodoulopoulos, Oana Cocarascu, and Arpit Mittal.
\newblock The fact extraction and {VER}ification over unstructured and structured information ({FEVEROUS}) shared task.
\newblock In Rami Aly, Christos Christodoulopoulos, Oana Cocarascu, Zhijiang Guo, Arpit Mittal, Michael Schlichtkrull, James Thorne, and Andreas Vlachos, editors, {\em Proceedings of the Fourth Workshop on Fact Extraction and VERification (FEVER)}, pages 1--13, Dominican Republic, November 2021. Association for Computational Linguistics.

\bibitem{banarjee2005}
Satanjeev Banerjee and Alon Lavie.
\newblock {METEOR}: An automatic metric for {MT} evaluation with improved correlation with human judgments.
\newblock In {\em Proceedings of the {ACL} Workshop on Intrinsic and Extrinsic Evaluation Measures for Machine Translation and/or Summarization}, pages 65--72, Ann Arbor, Michigan, June 2005. Association for Computational Linguistics.

\bibitem{blecher_nougat_2023}
Lukas Blecher, Guillem Cucurull, Thomas Scialom, and Robert Stojnic.
\newblock Nougat: {Neural} {Optical} {Understanding} for {Academic} {Documents}, August 2023.
\newblock arXiv:2308.13418 [cs].

\bibitem{brown2020language}
Tom Brown, Benjamin Mann, Nick Ryder, Melanie Subbiah, Jared~D Kaplan, Prafulla Dhariwal, Arvind Neelakantan, Pranav Shyam, Girish Sastry, Amanda Askell, et~al.
\newblock Language models are few-shot learners.
\newblock {\em Advances in Neural Information Processing Systems}, 33:1877--1901, 2020.

\bibitem{Chen2020TabFact:}
Wenhu Chen, Hongmin Wang, Jianshu Chen, Yunkai Zhang, Hong Wang, Shiyang Li, Xiyou Zhou, and William~Yang Wang.
\newblock Tabfact: A large-scale dataset for table-based fact verification.
\newblock In {\em International Conference on Learning Representations}, 2020.

\bibitem{chen-etal-2021-finqa}
Zhiyu Chen, Wenhu Chen, Charese Smiley, Sameena Shah, Iana Borova, Dylan Langdon, Reema Moussa, Matt Beane, Ting-Hao Huang, Bryan Routledge, and William~Yang Wang.
\newblock {F}in{QA}: A dataset of numerical reasoning over financial data.
\newblock In Marie-Francine Moens, Xuanjing Huang, Lucia Specia, and Scott Wen-tau Yih, editors, {\em Proceedings of the 2021 Conference on Empirical Methods in Natural Language Processing}, pages 3697--3711, Online and Punta Cana, Dominican Republic, November 2021. Association for Computational Linguistics.

\bibitem{cheng-etal-2022-hitab}
Zhoujun Cheng, Haoyu Dong, Zhiruo Wang, Ran Jia, Jiaqi Guo, Yan Gao, Shi Han, Jian-Guang Lou, and Dongmei Zhang.
\newblock {H}i{T}ab: A hierarchical table dataset for question answering and natural language generation.
\newblock In Smaranda Muresan, Preslav Nakov, and Aline Villavicencio, editors, {\em Proceedings of the 60th Annual Meeting of the Association for Computational Linguistics (Volume 1: Long Papers)}, pages 1094--1110, Dublin, Ireland, May 2022. Association for Computational Linguistics.

\bibitem{chiang-chen-2019-semantically}
Ting-Rui Chiang and Yun-Nung Chen.
\newblock Semantically-aligned equation generation for solving and reasoning math word problems.
\newblock In Jill Burstein, Christy Doran, and Thamar Solorio, editors, {\em Proceedings of the 2019 Conference of the North {A}merican Chapter of the Association for Computational Linguistics: Human Language Technologies, Volume 1 (Long and Short Papers)}, pages 2656--2668, Minneapolis, Minnesota, June 2019. Association for Computational Linguistics.

\bibitem{clark_extracting_nodate}
Christopher Clark and Santosh Divvala.
\newblock Extracting {Figures}, {Tables}, and {Captions} from {Computer} {Science} {Papers}.

\bibitem{das_visual_2017}
Abhishek Das, Satwik Kottur, Khushi Gupta, Avi Singh, Deshraj Yadav, José M.~F. Moura, Devi Parikh, and Dhruv Batra.
\newblock Visual {Dialog}.
\newblock {\em arXiv:1611.08669 [cs]}, August 2017.
\newblock arXiv: 1611.08669.

\bibitem{deng-etal-2022-pacific}
Yang Deng, Wenqiang Lei, Wenxuan Zhang, Wai Lam, and Tat-Seng Chua.
\newblock {PACIFIC}: Towards proactive conversational question answering over tabular and textual data in finance.
\newblock In Yoav Goldberg, Zornitsa Kozareva, and Yue Zhang, editors, {\em Proceedings of the 2022 Conference on Empirical Methods in Natural Language Processing}, pages 6970--6984, Abu Dhabi, United Arab Emirates, December 2022. Association for Computational Linguistics.

\bibitem{dettmers_qlora_2023}
Tim Dettmers, Artidoro Pagnoni, Ari Holtzman, and Luke Zettlemoyer.
\newblock {QLoRA}: {Efficient} {Finetuning} of {Quantized} {LLMs}.
\newblock In A.~Oh, T.~Neumann, A.~Globerson, K.~Saenko, M.~Hardt, and S.~Levine, editors, {\em Advances in {Neural} {Information} {Processing} {Systems}}, volume~36, pages 10088--10115. Curran Associates, Inc., 2023.

\bibitem{eisenschlos-etal-2020-understanding}
Julian Eisenschlos, Syrine Krichene, and Thomas M{\"u}ller.
\newblock Understanding tables with intermediate pre-training.
\newblock In Trevor Cohn, Yulan He, and Yang Liu, editors, {\em Findings of the Association for Computational Linguistics: EMNLP 2020}, pages 281--296, Online, November 2020. Association for Computational Linguistics.

\bibitem{fei_towards_2021}
Zhengcong Fei, Zekang Li, Jinchao Zhang, Yang Feng, and Jie Zhou.
\newblock Towards {Expressive} {Communication} with {Internet} {Memes}: {A} {New} {Multimodal} {Conversation} {Dataset} and {Benchmark}.
\newblock {\em arXiv:2109.01839 [cs]}, September 2021.
\newblock arXiv: 2109.01839.

\bibitem{gong-etal-2020-tablegpt}
Heng Gong, Yawei Sun, Xiaocheng Feng, Bing Qin, Wei Bi, Xiaojiang Liu, and Ting Liu.
\newblock {T}able{GPT}: Few-shot table-to-text generation with table structure reconstruction and content matching.
\newblock In Donia Scott, Nuria Bel, and Chengqing Zong, editors, {\em Proceedings of the 28th International Conference on Computational Linguistics}, pages 1978--1988, Barcelona, Spain (Online), December 2020. International Committee on Computational Linguistics.

\bibitem{gong_recognizing_2021}
Ming Gong, Xin Wei, Diane Oyen, Jian Wu, Martin Gryder, and Liping Yang.
\newblock Recognizing {Figure} {Labels} in {Patents}.
\newblock In {\em {SDU}@ {AAAI}}, 2021.

\bibitem{gu-etal-2022-pasta}
Zihui Gu, Ju~Fan, Nan Tang, Preslav Nakov, Xiaoman Zhao, and Xiaoyong Du.
\newblock {PASTA}: Table-operations aware fact verification via sentence-table cloze pre-training.
\newblock In Yoav Goldberg, Zornitsa Kozareva, and Yue Zhang, editors, {\em Proceedings of the 2022 Conference on Empirical Methods in Natural Language Processing}, pages 4971--4983, Abu Dhabi, United Arab Emirates, December 2022. Association for Computational Linguistics.

\bibitem{hannan_manymodalqa_2020}
Darryl Hannan, Akshay Jain, and Mohit Bansal.
\newblock {ManyModalQA}: {Modality} {Disambiguation} and {QA} over {Diverse} {Inputs}.
\newblock {\em arXiv:2001.08034 [cs]}, January 2020.
\newblock arXiv: 2001.08034.

\bibitem{hannan2020manymodalqa}
Darryl Hannan, Akshay Jain, and Mohit Bansal.
\newblock Manymodalqa: Modality disambiguation and qa over diverse inputs.
\newblock In {\em Proceedings of the AAAI Conference on Artificial Intelligence}, volume~34, pages 7879--7886, 2020.

\bibitem{herzig-etal-2021-open}
Jonathan Herzig, Thomas M{\"u}ller, Syrine Krichene, and Julian Eisenschlos.
\newblock Open domain question answering over tables via dense retrieval.
\newblock In Kristina Toutanova, Anna Rumshisky, Luke Zettlemoyer, Dilek Hakkani-Tur, Iz~Beltagy, Steven Bethard, Ryan Cotterell, Tanmoy Chakraborty, and Yichao Zhou, editors, {\em Proceedings of the 2021 Conference of the North American Chapter of the Association for Computational Linguistics: Human Language Technologies}, pages 512--519, Online, June 2021. Association for Computational Linguistics.

\bibitem{herzig-etal-2020-tapas}
Jonathan Herzig, Pawel~Krzysztof Nowak, Thomas M{\"u}ller, Francesco Piccinno, and Julian Eisenschlos.
\newblock {T}a{P}as: Weakly supervised table parsing via pre-training.
\newblock In Dan Jurafsky, Joyce Chai, Natalie Schluter, and Joel Tetreault, editors, {\em Proceedings of the 58th Annual Meeting of the Association for Computational Linguistics}, pages 4320--4333, Online, July 2020. Association for Computational Linguistics.

\bibitem{hsu-etal-2021-scicap-generating}
Ting-Yao Hsu, C~Lee Giles, and Ting-Hao Huang.
\newblock {S}ci{C}ap: Generating captions for scientific figures.
\newblock In Marie-Francine Moens, Xuanjing Huang, Lucia Specia, and Scott Wen-tau Yih, editors, {\em Findings of the Association for Computational Linguistics: EMNLP 2021}, pages 3258--3264, Punta Cana, Dominican Republic, November 2021. Association for Computational Linguistics.

\bibitem{jiang-etal-2023-structgpt}
Jinhao Jiang, Kun Zhou, Zican Dong, Keming Ye, Xin Zhao, and Ji-Rong Wen.
\newblock {S}truct{GPT}: A general framework for large language model to reason over structured data.
\newblock In Houda Bouamor, Juan Pino, and Kalika Bali, editors, {\em Proceedings of the 2023 Conference on Empirical Methods in Natural Language Processing}, pages 9237--9251, Singapore, December 2023. Association for Computational Linguistics.

\bibitem{kottur-moon-2023-overview}
Satwik Kottur and Seungwhan Moon.
\newblock Overview of situated and interactive multimodal conversations ({SIMMC}) 2.1 track at {DSTC} 11.
\newblock In Yun-Nung Chen, Paul Crook, Michel Galley, Sarik Ghazarian, Chulaka Gunasekara, Raghav Gupta, Behnam Hedayatnia, Satwik Kottur, Seungwhan Moon, and Chen Zhang, editors, {\em Proceedings of The Eleventh Dialog System Technology Challenge}, pages 235--241, Prague, Czech Republic, September 2023. Association for Computational Linguistics.

\bibitem{landhuis2016scientific}
Esther Landhuis.
\newblock Scientific literature: Information overload.
\newblock {\em Nature}, 535(7612):457--458, 2016.

\bibitem{lin_rouge_2004}
Chin-Yew Lin.
\newblock {ROUGE}: {A} {Package} for {Automatic} {Evaluation} of {Summaries}.
\newblock In {\em Text {Summarization} {Branches} {Out}}, pages 74--81, Barcelona, Spain, July 2004. Association for Computational Linguistics.

\bibitem{lin_microsoft_2014}
Tsung-Yi Lin, Michael Maire, Serge Belongie, James Hays, Pietro Perona, Deva Ramanan, Piotr Dollár, and C~Lawrence Zitnick.
\newblock Microsoft coco: {Common} objects in context.
\newblock In {\em Computer {Vision}–{ECCV} 2014: 13th {European} {Conference}, {Zurich}, {Switzerland}, {September} 6-12, 2014, {Proceedings}, {Part} {V} 13}, pages 740--755. Springer, 2014.

\bibitem{liu2021tapex}
Qian Liu, Bei Chen, Jiaqi Guo, Morteza Ziyadi, Zeqi Lin, Weizhu Chen, and Jian-Guang Lou.
\newblock Tapex: Table pre-training via learning a neural sql executor.
\newblock {\em arXiv preprint arXiv:2107.07653}, 2021.

\bibitem{marino_ok-vqa_2019}
Kenneth Marino, Mohammad Rastegari, Ali Farhadi, and Roozbeh Mottaghi.
\newblock {OK}-{VQA}: {A} {Visual} {Question} {Answering} {Benchmark} {Requiring} {External} {Knowledge}.
\newblock In {\em Conference on {Computer} {Vision} and {Pattern} {Recognition} ({CVPR})}, 2019.

\bibitem{SIMMC:2020}
Seungwhan Moon, Satwik Kottur, Paul~A. Crook, Ankita De, Shivani Poddar, Theodore Levin, David Whitney, Daniel Difranco, Ahmad Beirami, Eunjoon Cho, Rajen Subba, and Alborz Geramifard.
\newblock Situated and interactive multimodal conversations.
\newblock {\em CoRR}, abs/2006.01460, 2020.

\bibitem{mostafazadeh_image-grounded_2017}
Nasrin Mostafazadeh, Chris Brockett, Bill Dolan, Michel Galley, Jianfeng Gao, Georgios~P. Spithourakis, and Lucy Vanderwende.
\newblock Image-{Grounded} {Conversations}: {Multimodal} {Context} for {Natural} {Question} and {Response} {Generation}.
\newblock {\em arXiv:1701.08251 [cs]}, April 2017.
\newblock arXiv: 1701.08251.

\bibitem{nakamura-etal-2022-hybridialogue}
Kai Nakamura, Sharon Levy, Yi-Lin Tuan, Wenhu Chen, and William~Yang Wang.
\newblock {H}ybri{D}ialogue: An information-seeking dialogue dataset grounded on tabular and textual data.
\newblock In Smaranda Muresan, Preslav Nakov, and Aline Villavicencio, editors, {\em Findings of the Association for Computational Linguistics: ACL 2022}, pages 481--492, Dublin, Ireland, May 2022. Association for Computational Linguistics.

\bibitem{nan-etal-2022-fetaqa}
Linyong Nan, Chiachun Hsieh, Ziming Mao, Xi~Victoria Lin, Neha Verma, Rui Zhang, Wojciech Kry{\'s}ci{\'n}ski, Hailey Schoelkopf, Riley Kong, Xiangru Tang, Mutethia Mutuma, Ben Rosand, Isabel Trindade, Renusree Bandaru, Jacob Cunningham, Caiming Xiong, Dragomir Radev, and Dragomir Radev.
\newblock {F}e{T}a{QA}: Free-form table question answering.
\newblock {\em Transactions of the Association for Computational Linguistics}, 10:35--49, 2022.

\bibitem{pasupat-liang-2015-compositional}
Panupong Pasupat and Percy Liang.
\newblock Compositional semantic parsing on semi-structured tables.
\newblock In Chengqing Zong and Michael Strube, editors, {\em Proceedings of the 53rd Annual Meeting of the Association for Computational Linguistics and the 7th International Joint Conference on Natural Language Processing (Volume 1: Long Papers)}, pages 1470--1480, Beijing, China, July 2015. Association for Computational Linguistics.

\bibitem{peng2021mathbert}
Shuai Peng, Ke~Yuan, Liangcai Gao, and Zhi Tang.
\newblock Mathbert: A pre-trained model for mathematical formula understanding.
\newblock {\em arXiv preprint arXiv:2105.00377}, 2021.

\bibitem{radford2019language}
Alec Radford, Jeffrey Wu, Rewon Child, David Luan, Dario Amodei, Ilya Sutskever, et~al.
\newblock Language models are unsupervised multitask learners.
\newblock {\em OpenAI blog}, 1(8):9, 2019.

\bibitem{ramesh-etal-2023-comparative}
Krithika Ramesh, Arnav Chavan, Shrey Pandit, and Sunayana Sitaram.
\newblock A comparative study on the impact of model compression techniques on fairness in language models.
\newblock In Anna Rogers, Jordan Boyd-Graber, and Naoaki Okazaki, editors, {\em Proceedings of the 61st Annual Meeting of the Association for Computational Linguistics (Volume 1: Long Papers)}, pages 15762--15782, Toronto, Canada, July 2023. Association for Computational Linguistics.

\bibitem{reichman_outside_2023}
Benjamin~Z Reichman, Anirudh Sundar, Christopher Richardson, Tamara Zubatiy, Prithwijit Chowdhury, Aaryan Shah, Jack Truxal, Micah Grimes, Dristi Shah, Woo~Ju Chee, and {others}.
\newblock Outside knowledge visual question answering version 2.0.
\newblock In {\em {ICASSP} 2023-2023 {IEEE} {International} {Conference} on {Acoustics}, {Speech} and {Signal} {Processing} ({ICASSP})}, pages 1--5. IEEE, 2023.

\bibitem{sarkar2023testing}
Soumajyoti Sarkar and Leonard Lausen.
\newblock Testing the limits of unified sequence to sequence llm pretraining on diverse table data tasks, 2023.

\bibitem{scarlatos-lan-2023-tree}
Alexander Scarlatos and Andrew Lan.
\newblock Tree-based representation and generation of natural and mathematical language.
\newblock In Anna Rogers, Jordan Boyd-Graber, and Naoaki Okazaki, editors, {\em Proceedings of the 61st Annual Meeting of the Association for Computational Linguistics (Volume 1: Long Papers)}, pages 3714--3730, Toronto, Canada, July 2023. Association for Computational Linguistics.

\bibitem{shuster_image_2020}
Kurt Shuster, Samuel Humeau, Antoine Bordes, and Jason Weston.
\newblock Image {Chat}: {Engaging} {Grounded} {Conversations}.
\newblock {\em arXiv:1811.00945 [cs]}, April 2020.
\newblock arXiv: 1811.00945.

\bibitem{10.1145/3616855.3635752}
Yuan Sui, Mengyu Zhou, Mingjie Zhou, Shi Han, and Dongmei Zhang.
\newblock Table meets llm: Can large language models understand structured table data? a benchmark and empirical study.
\newblock In {\em Proceedings of the 17th ACM International Conference on Web Search and Data Mining}, WSDM '24, page 645–654, New York, NY, USA, 2024. Association for Computing Machinery.

\bibitem{sundar-heck-2022-multimodal}
Anirudh Sundar and Larry Heck.
\newblock Multimodal conversational {AI}: A survey of datasets and approaches.
\newblock In Bing Liu, Alexandros Papangelis, Stefan Ultes, Abhinav Rastogi, Yun-Nung Chen, Georgios Spithourakis, Elnaz Nouri, and Weiyan Shi, editors, {\em Proceedings of the 4th Workshop on NLP for Conversational AI}, pages 131--147, Dublin, Ireland, May 2022. Association for Computational Linguistics.

\bibitem{sundar2024itbls}
Anirudh Sundar, Christopher Richardson, William Gay, and Larry Heck.
\newblock itbls: A dataset of interactive conversations over tabular information.
\newblock {\em arXiv preprint arXiv:2404.12580}, 2024.

\bibitem{sundar-heck-2023-ctbls}
Anirudh~S. Sundar and Larry Heck.
\newblock c{TBLS}: Augmenting large language models with conversational tables.
\newblock In Yun-Nung Chen and Abhinav Rastogi, editors, {\em Proceedings of the 5th Workshop on NLP for Conversational AI (NLP4ConvAI 2023)}, pages 59--70, Toronto, Canada, July 2023. Association for Computational Linguistics.

\bibitem{tan_scientific_2022}
Hao Tan, Chen-Tse Tsai, Yujie He, and Mohit Bansal.
\newblock Scientific {Chart} {Summarization}: {Datasets} and {Improved} {Text} {Modeling}.
\newblock In {\em {SDU}@{AAAI}}, 2022.

\bibitem{thomee_yfcc100m_2016}
Bart Thomee, David~A. Shamma, Gerald Friedland, Benjamin Elizalde, Karl Ni, Douglas Poland, Damian Borth, and Li-Jia Li.
\newblock {YFCC100M}: {The} {New} {Data} in {Multimedia} {Research}.
\newblock {\em Communications of the ACM}, 59(2):64--73, January 2016.
\newblock arXiv: 1503.01817.

\bibitem{touvron2023llama}
Hugo Touvron, Thibaut Lavril, Gautier Izacard, Xavier Martinet, Marie-Anne Lachaux, Timoth{\'e}e Lacroix, Baptiste Rozi{\`e}re, Naman Goyal, Eric Hambro, Faisal Azhar, et~al.
\newblock Llama: Open and efficient foundation language models.
\newblock {\em arXiv preprint arXiv:2302.13971}, 2023.

\bibitem{wang-etal-2021-math}
Zichao Wang, Andrew Lan, and Richard Baraniuk.
\newblock Math word problem generation with mathematical consistency and problem context constraints.
\newblock In Marie-Francine Moens, Xuanjing Huang, Lucia Specia, and Scott Wen-tau Yih, editors, {\em Proceedings of the 2021 Conference on Empirical Methods in Natural Language Processing}, pages 5986--5999, Online and Punta Cana, Dominican Republic, November 2021. Association for Computational Linguistics.

\bibitem{wang2021scientific}
Zichao Wang, Mengxue Zhang, Richard~G Baraniuk, and Andrew~S Lan.
\newblock Scientific formula retrieval via tree embeddings.
\newblock In {\em 2021 IEEE International Conference on Big Data (Big Data)}, pages 1493--1503. IEEE, 2021.

\bibitem{yang-etal-2022-tableformer}
Jingfeng Yang, Aditya Gupta, Shyam Upadhyay, Luheng He, Rahul Goel, and Shachi Paul.
\newblock {T}able{F}ormer: Robust transformer modeling for table-text encoding.
\newblock In Smaranda Muresan, Preslav Nakov, and Aline Villavicencio, editors, {\em Proceedings of the 60th Annual Meeting of the Association for Computational Linguistics (Volume 1: Long Papers)}, pages 528--537, Dublin, Ireland, May 2022. Association for Computational Linguistics.

\bibitem{yasunaga2019topiceq}
Michihiro Yasunaga and John~D Lafferty.
\newblock Topiceq: A joint topic and mathematical equation model for scientific texts.
\newblock In {\em Proceedings of the AAAI conference on artificial intelligence}, volume~33, pages 7394--7401, 2019.

\bibitem{yin-etal-2020-tabert}
Pengcheng Yin, Graham Neubig, Wen-tau Yih, and Sebastian Riedel.
\newblock {T}a{BERT}: Pretraining for joint understanding of textual and tabular data.
\newblock In Dan Jurafsky, Joyce Chai, Natalie Schluter, and Joel Tetreault, editors, {\em Proceedings of the 58th Annual Meeting of the Association for Computational Linguistics}, pages 8413--8426, Online, July 2020. Association for Computational Linguistics.

\bibitem{zha2023tablegpt}
Liangyu Zha, Junlin Zhou, Liyao Li, Rui Wang, Qingyi Huang, Saisai Yang, Jing Yuan, Changbao Su, Xiang Li, Aofeng Su, et~al.
\newblock Tablegpt: Towards unifying tables, nature language and commands into one gpt.
\newblock {\em arXiv preprint arXiv:2307.08674}, 2023.

\bibitem{zhang2023tablellama}
Tianshu Zhang, Xiang Yue, Yifei Li, and Huan Sun.
\newblock Tablellama: Towards open large generalist models for tables.
\newblock {\em arXiv preprint arXiv:2311.09206}, 2023.

\bibitem{zhang_bertscore_2020}
Tianyi Zhang, Varsha Kishore*, Felix Wu*, Kilian~Q. Weinberger, and Yoav Artzi.
\newblock {BERTScore}: {Evaluating} {Text} {Generation} with {BERT}.
\newblock In {\em International {Conference} on {Learning} {Representations}}, 2020.

\bibitem{zhao-etal-2022-multihiertt}
Yilun Zhao, Yunxiang Li, Chenying Li, and Rui Zhang.
\newblock {M}ulti{H}iertt: Numerical reasoning over multi hierarchical tabular and textual data.
\newblock In Smaranda Muresan, Preslav Nakov, and Aline Villavicencio, editors, {\em Proceedings of the 60th Annual Meeting of the Association for Computational Linguistics (Volume 1: Long Papers)}, pages 6588--6600, Dublin, Ireland, May 2022. Association for Computational Linguistics.

\bibitem{zhu-etal-2021-tat}
Fengbin Zhu, Wenqiang Lei, Youcheng Huang, Chao Wang, Shuo Zhang, Jiancheng Lv, Fuli Feng, and Tat-Seng Chua.
\newblock {TAT}-{QA}: A question answering benchmark on a hybrid of tabular and textual content in finance.
\newblock In Chengqing Zong, Fei Xia, Wenjie Li, and Roberto Navigli, editors, {\em Proceedings of the 59th Annual Meeting of the Association for Computational Linguistics and the 11th International Joint Conference on Natural Language Processing (Volume 1: Long Papers)}, pages 3277--3287, Online, August 2021. Association for Computational Linguistics.

\bibitem{zhu_visual7w_2016}
Yuke Zhu, Oliver Groth, Michael Bernstein, and Li~Fei-Fei.
\newblock {Visual7W}: {Grounded} {Question} {Answering} in {Images}.
\newblock {\em arXiv:1511.03416 [cs]}, April 2016.
\newblock arXiv: 1511.03416.

\end{thebibliography}

\end{document}